\documentclass[letterpaper, 10 pt, conference]{ieeeconf}  % Comment this line out if you need a4paper

\IEEEoverridecommandlockouts                              % This command is only needed if 
\usepackage{cite}
\usepackage{amsmath,amssymb,amsfonts}
\usepackage{amssymb}% http://ctan.org/pkg/amssymb
\usepackage{pifont}% http://ctan.org/pkg/pifont
\usepackage{gensymb}
\usepackage[ruled]{algorithm2e}
\usepackage{algpseudocode}
\usepackage{graphicx}
\usepackage{textcomp}
\usepackage{siunitx}
\usepackage{multirow}
\usepackage{bbding}
\usepackage{booktabs}
\usepackage{lscape}
\usepackage{tablefootnote}
\usepackage{refcount}
\usepackage{threeparttablex}
\usepackage{kotex}
\usepackage{color, soul}
\usepackage[table,xcdraw]{xcolor}
\usepackage{hyperref}

\SetCommentSty{mycommfont}

\newcolumntype{?}[1]{!{\vrule width #1}}

\newcommand{\etal}{\textit{et al}. }
\newcommand*{\rom}[1]{\uppercase\expandafter{\romannumeral #1\relax}}
\newcolumntype{M}[1]{>{\centering\arraybackslash}m{#1}}

\title{\LARGE \bf Radar-Based NLoS Pedestrian Localization for Darting-Out Scenarios Near Parked Vehicles with Camera-Assisted Point Cloud Interpretation}

\author{Hee-Yeun Kim$^{1}$, Byeonggyu Park$^{1}$, Byonghyok Choi$^{2}$, Hansang Cho$^{2}$, Byungkwan Kim$^{3}$,\\Soomok Lee$^{4}$, Mingu Jeon$^{1}$, Seung-Woo Seo$^{1}$, and Seong-Woo Kim$^{1}$% <-this % stops a space\\%Byungkwan Kim$^{5}$, Soomok Lee$^{6}$,
	\thanks{This work was supported by Samsung Electro-Mechanics Co., Ltd., the National Research Foundation of Korea (NRF) through the Ministry of Science and ICT under Grant 2021R1A2C1093957, Korea Institute for Advancement of Technology (KIAT) grant funded by the Korea Government (MOTIE) (P0020536, HRD Program for Industrial Innovation), the Korean Ministry of Land, Infrastructure and Transport (MOLIT) as the Innovative Talent Education Program for Smart City, and by the Institute of Engineering Research at Seoul National University, which provided the research facilities for this work.}
	\thanks{$^{1}$Seoul National University, $^{2}$Samsung Electro-Mechanics Co., Ltd., $^{3}$Chungnam National University, $^{4}$Ajou University. Correspondence to: Mingu Jeon, Seung-Woo Seo, and Seong-Woo Kim {\tt\small \{mingujeon; sseo; snwoo\}@snu.ac.kr}}%
}

\begin{document}

\maketitle
\thispagestyle{empty}
\pagestyle{empty}

%%%%%%%%%%%%%%%%%%%%%%%%%%%%%%%%%%%%%%%%%%%%%%%%%%%%%%%%%%%%%%%%%%%%%%%%%%%%%%%%
\begin{abstract}
The presence of Non-Line-of-Sight (NLoS) blind spots resulting from roadside parking in urban environments poses a significant challenge to road safety, particularly due to the sudden emergence of pedestrians. mmWave technology leverages diffraction and reflection to observe NLoS regions, and recent studies have demonstrated its potential for detecting obscured objects. However, existing approaches predominantly rely on predefined spatial information or assume simple wall reflections, thereby limiting their generalizability and practical applicability. A particular challenge arises in scenarios where pedestrians suddenly appear from between parked vehicles, as these parked vehicles act as temporary spatial obstructions. Furthermore, since parked vehicles are dynamic and may relocate over time, spatial information obtained from satellite maps or other predefined sources may not accurately reflect real-time road conditions, leading to erroneous sensor interpretations. To address this limitation, we propose an NLoS pedestrian localization framework that integrates monocular camera image with 2D radar point cloud (PCD) data. The proposed method initially detects parked vehicles through image segmentation, estimates depth to infer approximate spatial characteristics, and subsequently refines this information using 2D radar PCD to achieve precise spatial inference. Experimental evaluations conducted in real-world urban road environments demonstrate that the proposed approach enhances early pedestrian detection and contributes to improved road safety. Supplementary materials are available at https://hiyeun.github.io/NLoS/.

% 도심 환경에서 길가 주차로 인해 발생하는 비가시선(NLoS) 사각지대의 존재는 특히 보행자의 갑작스러운 출현으로 인해 도로 안전에 심각한 문제를 야기합니다. 밀리미터파(mmWave) 기술은 회절과 반사를 활용하여 NLoS 영역을 관찰하며, 최근 연구들은 가려진 물체를 감지하는 데 잠재력을 보여주었습니다. 그러나 기존 접근 방식은 주로 미리 정의된 공간 정보에 의존하거나 단순한 벽 반사를 가정하므로 일반화 가능성과 실제 적용 가능성이 제한됩니다. 특히 주차된 차량 사이에서 보행자가 갑자기 나타나는 시나리오에서 어려움이 발생하는데, 이러한 차량은 일시적인 공간 장애물 역할을 하기 때문입니다. 또한 주차된 차량은 역동적이고 시간이 지남에 따라 위치가 변경될 수 있으므로 위성 지도 또는 기타 미리 정의된 소스에서 얻은 공간 정보가 실시간 도로 상황을 정확하게 반영하지 못하여 잘못된 센서 해석을 초래할 수 있습니다. 이러한 제한을 해결하기 위해 우리는 단안 카메라 이미지와 2D 레이더 포인트 클라우드(PCD) 데이터를 통합하는 NLoS 보행자 위치 추정 프레임워크를 제안합니다. 제안된 방법은 먼저 이미지 분할을 통해 주차된 차량을 감지하고, 깊이를 추정하여 대략적인 공간 특성을 유추한 다음, 2D 레이더 PCD를 사용하여 이 정보를 개선하여 정확한 공간 추론을 달성합니다. 실제 도시 도로 환경에서 수행된 실험 평가는 제안된 접근 방식이 초기 보행자 감지를 향상시키고 도로 안전 개선에 기여함을 보여줍니다. 
\end{abstract}

\begin{keywords}
2D radar point cloud, darting out, monocular depth estimation, non-line-of-sight, parked vehicle.
\end{keywords}

\section{Introduction}\label{sec:Introduction}
\begin{figure}[t!]
	\centering
	\centering
	\includegraphics[width=\linewidth]{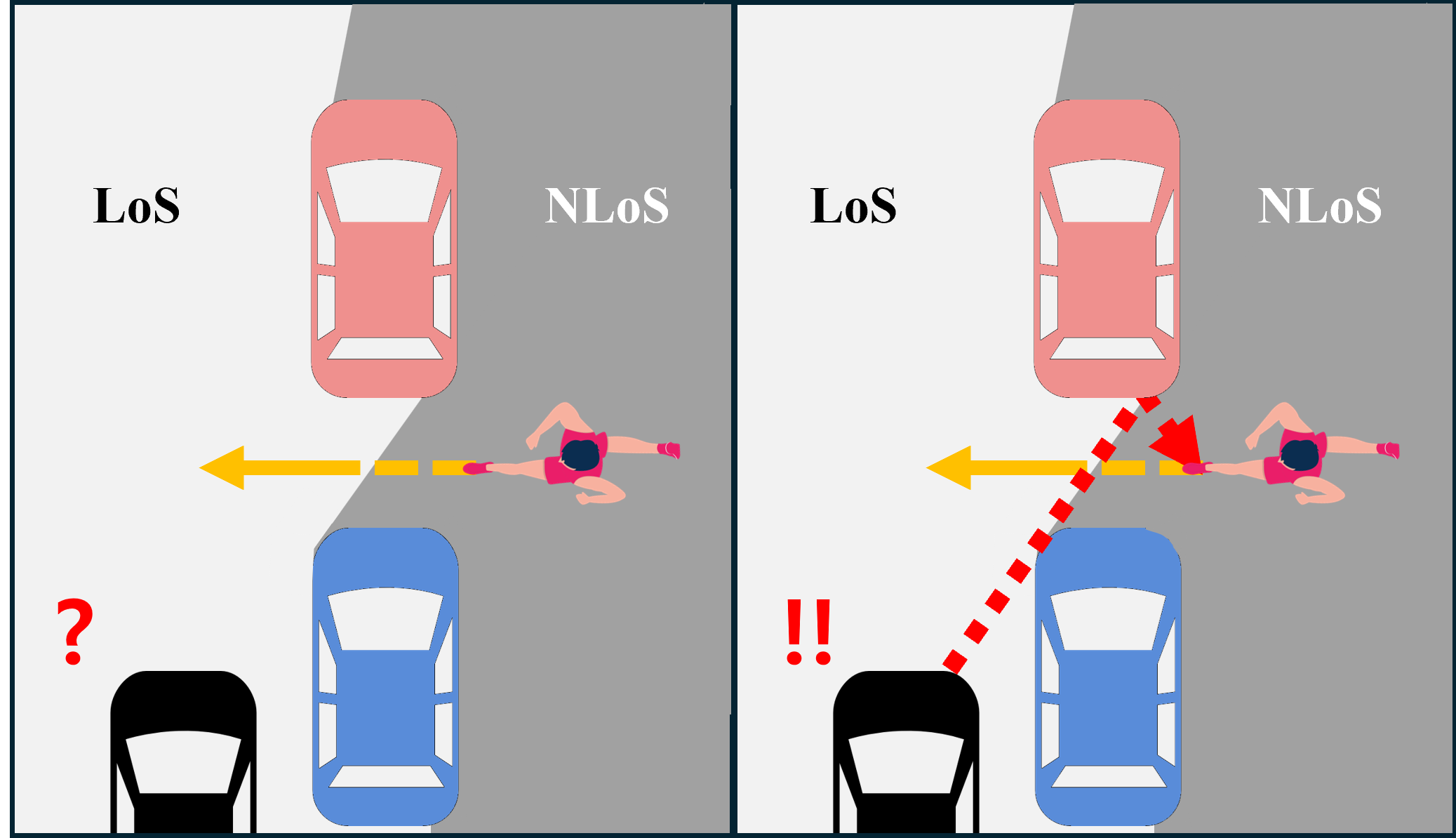}
	% \vspace{-0.7cm}
	\caption{Illustration of the target scenario. (Left) The darting out pedestrian cannot be observed at driver's Field-of-view. (Right) The darting out pedestrian can be observed by using the proposed method. The red line means the radar signal path.}
	\label{fig:scenario}
\end{figure}

In urban environments, roadside parking is a common occurrence. However, the blind spots created between parked vehicles pose significant safety risks and can be a great threat to drivers. When navigating road with evident blind spot, such as corners, drivers instinctively reduce speed. However, on straight roads with frequent roadside parking, drivers often assume they have a clear sight and may not decelerate. In reality, jaywalking pedestrians frequently emerge from between parked vehicles, and particularly, children may exhibit sudden and unpredictable behavior, such as running into the road without checking their surroundings, as shown in Figure \ref{fig:scenario}. If a driver begins to decelerate after visually detecting the pedestrian, the reaction time may be insufficient, potentially leading to severe accidents.

Observing and localizing pedestrians in Non-Line-of-Sight (NLoS) regions can help mitigate such accidents, and V2X communication-based driving environment awareness is a widely researched in this context \cite{rebsamen2012utilizing}. For instance, several studies have proposed cooperative driving strategies in which V2V communication enables the lead vehicle to share environmental perception information with following vehicles, improving both accident avoidance and overall traffic flow \cite{kim2013cooperative, liu2013motion, kim2014multivehicle}. However, V2X communication has a limitation in that it cannot be used at all in situations where infrastructure does not exist, and the cost of continuous maintenance is high.

These limitations highlight the need for techniques to quantify the risks in the NLoS domain of driving vehicles themselves. mmWave radar and sound-based sensing have the advantage of detecting occluded objects due to their diffraction and reflective properties, and recent studies have demonstrated the feasibility of NLoS detection using these signals\cite{jeon2024non, zhu2023non, wang2021multipath, chen2022non, luo2025reflective, shen2023darting, palffy2022detecting}. However, existing research relies on predefined spatial information or assumes simple reflective surfaces such as walls, which constrains their applicability in real-world scenarios, such as pedestrians suddenly emerging from between parked vehicles.

The inability to independently estimate spatial information becomes a particularly critical issue when utilizing reflective waves. While the reflectivity of mmWave and sound enables the observation of NLoS regions, it simultaneously provides distorted observation data\cite{kraus2021radar}. Therefore, it is necessary to analyze the reflection paths of reflective waves based on spatial information to rectify these distortions. 

For instance, in scenarios where an NLoS object suddenly appears between parked vehicles, as illustrated in Figure \ref{fig:scenario}, the parked vehicle acts as a temporary spatial component. Consequently, spatial inference must incorporate the spatial state of these parked vehicles. However, since parked vehicles can move at any time, pre-defined spatial information obtained from sources such as satellite maps may not accurately reflect the real driving environment, leading to potential misinterpretations of observed data\cite{woo2024no}.

To address these challenges, this paper proposes a method that utilizes monocular camera images and 2D radar point cloud data (PCD) to infer spatial information including vehicles, and subsequently localize NLoS objects. The 2D radar PCD provides precise distance measurements and valuable observations of NLoS regions; however, its inherently low resolution results in sparse data, making it challenging to construct spatial information. Conversely, camera images contain rich features about the surroundings but suffer from reduced reliability in distance estimation due to variations in lighting conditions and surface textures\cite{ranftl2020towards}.

Leveraging these complementary characteristics, the proposed method employs an image segmentation model on the ego vehicle’s front camera image to detect objects classified as vehicles. A depth estimation approach is then used to obtain a coarse but approximate localization of these vehicles. This initial estimate is subsequently refined using radar-based distance correction, allowing the inclusion of reflective surfaces of vehicles that are not directly discernible in the sparse radar PCD. Using the spatial information derived in this manner, the reflection paths of radar PCD measurements are analyzed, enabling the early detection of potential threats emerging from blind spots, such as spaces between parked vehicles.
The main contributions of this paper can be summarized as follows:
\begin{itemize}
\item A novel image-based radar PCD interpretation pipeline for localizing darting-out NLoS pedestrians unexpectedly appears between parked vehicles in a driving environment where parked vehicles are used as reflectors is proposed.
\item Inference method of spatial information by interpreting sparse radar PCD using depth information extracted from images with imprecise distance measurements is proposed.
\item To demonstrate the applicability of proposed method in actual driving, the proposed method is validated using data collected in real-world, real-scale outdoor road environments.
\end{itemize}

\section{Related Works}\label{sec:RelatedWorks}

\begin{table}[t!]
	\centering
	\caption{Comparison of NLoS Ped. Localization Method Using Radar.}
	\label{tab:relatedworks}
	\resizebox{\columnwidth}{!}{
			\begin{tabular}{c|c|cc|cc}
					\hline
					\multirow{2}{*}{Methods}  & Input & \multicolumn{2}{c|}{Object}  & Sensor & Reflector \\
					\cline{3-4}
					 & type & Dynamic & Multi & fusion &  inference \\  \hline\hline
					
					\begin{tabular}[c]{@{}c@{}}Chen \etal \\\cite{chen2022non}\end{tabular}  & Signal &  & \checkmark &  & \checkmark  \\ 
					\hline
					
					\begin{tabular}[c]{@{}c@{}}Shen \etal \\\cite{shen2023darting}\end{tabular} & Signal & \checkmark &  &  & \checkmark \\
					\hline
									
 					\begin{tabular}[c]{@{}c@{}}Palffy \etal \\\cite{palffy2022detecting}\end{tabular}& \begin{tabular}[c]{@{}c@{}}PCD, \\ Image\end{tabular} & \checkmark &   & \checkmark & \\ 
					\midrule[0.9pt]
					
					\begin{tabular}[c]{@{}c@{}}\textbf{Proposed} \\\textbf{method}\end{tabular}& \begin{tabular}[c]{@{}c@{}}PCD, \\ Image\end{tabular} & \checkmark & \checkmark & \checkmark & \checkmark \\ 
					
					\hline
			\end{tabular}
		}
\end{table}

\subsection{Depth estimation using monocular camera}
Depth estimation, which predicts the distance to objects from a monocular camera image using learning-based approaches, has been extensively studied \cite{oquab2023dinov2, ranftl2020towards, Bhat_2021_CVPR}. Depth Anything V2 \cite{yang2025depth} has demonstrated significant performance improvements by leveraging large-scale multi-domain training on a Vision Transformer model, in addition to utilizing existing datasets collected in real-world environments. However, vision-based sensors inherently suffer from high sensitivity to lighting conditions, which remains a major limitation.

While depth estimation enables depth information retrieval without requiring active sensors like LiDAR, it does not provide direct distance measurements but rather estimates based on predictions, which results in lower accuracy and potential errors. Particularly in untrained environments, performance may degrade, and the method is highly susceptible to illumination conditions such as reflections and shadows, making it difficult to achieve stable and reliable results.Therefore, in order to obtain more accurate information, it is necessary to combine depth information with other sensor information.

\subsection{Reflective wave based NLoS object detection}
NLoS object detection has been extensively studied using highly reflective modalities \cite{zhu2023non, wang2021multipath}. Chen \etal proposed a method for multi-NLoS object detection in indoor L-shaped NLoS environments, utilizing Multiple-Input Multiple-Output (MIMO) radar, where multiple reflection path analysis and Time of Arrivals estimation were employed \cite{chen2022non}. However, this study is limited to indoor L-shaped corner scenarios, making it difficult to generalize to road environments, as summarized in Table \ref{tab:relatedworks}.

Palffy \etal quantified the risk of pedestrians darting into the road by using stereo cameras and radar to detect partially occluded pedestrians whose heads are visible behind a vehicle \cite{palffy2022detecting}. However, their method assumes that the pedestrian is partially occluded rather than fully occluded, significantly reducing both detection time and range. Additionally, while the vehicle itself is detected, it is only used for radar PCD filtering, failing to leverage the full potential of radar information for spatial reasoning.

Shen \etal introduced a method to detect darting-out pedestrians emerging from behind parked vehicles in urban environments by utilizing ground reflections in 3D MIMO mmWave radar \cite{shen2023darting}. However, this approach assumes that only a single vehicle is present, ensuring sufficient space behind it. It specifically addresses scenarios where pedestrians appear directly behind the vehicle rather than between two vehicles, which limits its practical applicability.

Luo \etal proposed a dictionary-based approach that leverages the geometric relationship between ghost targets generated by multipath reflections and their reflection points on surfaces. This method constructs a database of expected ghost target locations for each multipath ellipse reflection point \cite{luo2025reflective}. However, the computational complexity of this approach is extremely high, as it requires calculating the relationships between all possible multipath ellipses and ghost targets. To reduce this computational burden, all potential reflection conditions must be pre-defined in advance.

Consequently, existing NLoS localization methods utilizing mmWave are highly affected by the accuracy of spatial information inference. To tackle these limitations, it is crucial for a moving vehicle to infer spatial information and identify reflective surfaces when pre-defined spatial information is unavailable.

\section{Target Scenario and Problem Definition}\label{sec:ProblemDefinition}

The target scenario aims to detect darting-out pedestrians emerging from blind spots between parked vehicles while driving in road environments with a high density of roadside-parked vehicles. If a pedestrian approaching from an NLoS region is detected only after entering the Line-of-Sight (LoS) area while the vehicle is driving at full speed without deceleration, the available reaction time is insufficient for effective braking. Thus, early detection in the NLoS region is essential for preventing accidents.

To detect the darting-out pedestrians, the approximate distance information $D$ is estimated from ego-vehicle's front image $I$. Using the class information extracted from the image and the corresponding $D$ values, static radar PCD $R_s$ is analyzed to infer the spatial state of the surrounding vehicles, thereby estimating the spatial information $S$:
\begin{align}
    S = g(R_s \mid D),
\end{align}
where $g$ is the function for the spatial alignment. 

Subsequently, the predicted pedestrian location $X_{pred}$ based on $S$ can be estimated by analyzing the dynamic radar PCD $R_d$:
\begin{align}
    X_{pred}=h(R_d \mid S), 
\end{align}
where $h$ is the function for NLoS pedestrian localization.

To accurately assess the accident situation, the precision of the predicted NLoS pedestrian location is crucial. Since minimizing the localization error between $X_{pred}$ and the ground truth pedestrian position $X_{GT}$ is critical, the target problem is formulated as follows:
\begin{align}
\label{eq:3}
    X^*_{\text{pred}} = \arg \min_{X_{\text{pred}}} |X_{\text{pred}} - X_{\text{GT}}|,
\end{align}
where, $|\cdot |$ means the absolute distance.

% This paper focuses on accidents caused by the inability to detect and localize pedestrians moving in the NLoS region of a T-junction. Human drivers require a certain reaction time and distance, making accidents inevitable. To prevent such accidents, it is essential for an Active Emergency Braking System to actively intervene by minimizing the reaction distance through collision probability assessment.

% Therefore, the primary objective of this paper is to accurately predict the location $X_{pred}$ of NLoS pedestrians, as defined by the following equation:
% \[
% X_{pred}^*=\argmin_{X_{pred}}|X_{pred}-X_{GT}|,
% \]
% where $X_{GT}$ represents the ground truth location of the NLoS pedestrian, and $|\cdot|$ denotes the absolute error, representing the distance between $X_{pred}$ and $X_{GT}$. This will be discussed in detail in Section \ref{sec:Experiments}.

\section{Analysis Pipeline of 2D Radar Point Cloud\label{sec:ProposedMethod}}

\begin{figure}[t!]
	\centering
	\centering
	\includegraphics[width=\linewidth]{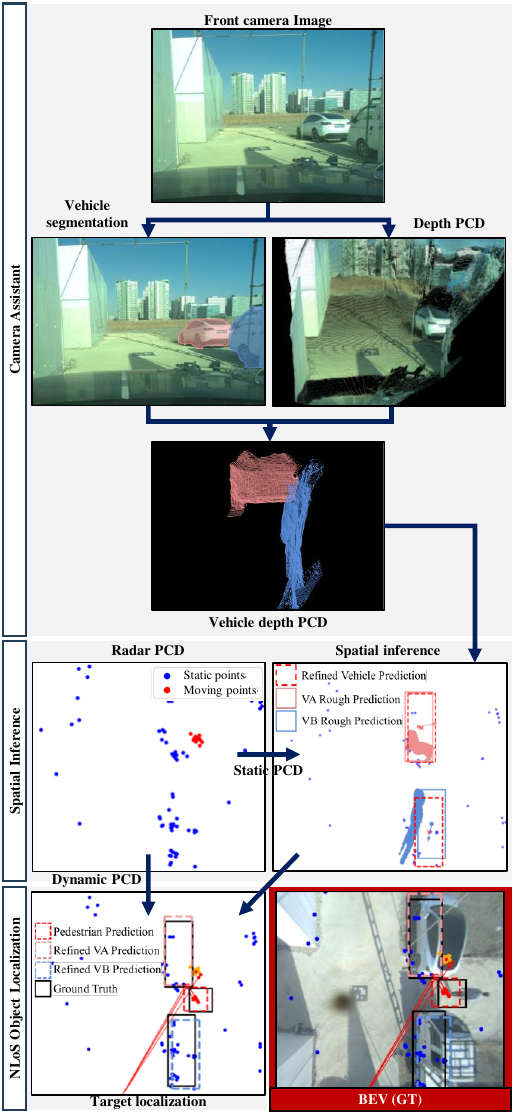}
        % \vspace{-0.7cm}
	\caption{The overall architecture and result for spatial inference and NLoS object localization.}
	\label{fig:flowchart}
\end{figure}

The data acquired through 2D mmWave radar exhibits high reflectivity, allowing observation of the surroundings, including NLoS regions. The extracted raw 2D radar PCD $R$ consists of both $R_s$ and $R_d$, as formulated below:
\begin{align}
        R &= R_s \cup R_d.
\end{align}

Static objects do not pose any potential risk to appear in the ego vehicle's trajectory; instead, they merely serve as sources of mmWave reflections. Hence, static objects can be regarded as spatial structures or reflectors. Consequently, all $R_s$ are considered spatial candidates, yet they inherently contain observations related to spatial structures $R_{reflector}$ as well as static noise $R_{s, noise}$, defined as follows:
\begin{align}
            R_s  &= R_{reflector} \cup R_{s,noise}.
\end{align}

Conversely, moving objects may enter the ego vehicle's trajectory from an NLoS region, posing potential risks. Thus, the primary target of interest for a driver is dynamic obejects. While all $R_d$ are target candidates, they also contain both obeservations of actual moving objects $R_{target}$ and dynamic noise $R_{d, noise}$, formulated as follows: 
\begin{align}
        R_d &= R_{target} \cup R_{d,noise}.
\end{align}

To identify NLoS targets, it is essential to filter out irrelevant data while retaining the desired information. Specifically, $R_{target}$ should be strictly composed of first-order reflection paths that accurately capture NLoS object information, eliminating clutter caused by noise or multiple reflections\cite{kraus2020using}. Additionally, to correctly interpret the first-order reflection path observed in $R_{target}$, it is necessary to analyze $R_s$ and extract $R_{reflector}$. 

% Therefore, for effective NLoS pedestrian localization, it is essential to remove \( R_{s,noise} \) from \( R_s \) to extract \( R_{reflector} \), which serves as the basis for estimating the reflection path and subsequently determining the actual location of \( R_d \). Furthermore, to accurately identify \( R_{target} \), it is necessary to eliminate \( R_{d,noise} \) from \( R_d \). 이 문제 해결을 위해 제안하는 overall architecture는 그림 xxx에 있습니다.
Therefore, for effective NLoS pedestrian localization, it is essential to remove \( R_{s,noise} \) from \( R_s \) to extract \( R_{reflector} \), which serves as the basis for estimating the reflection path and subsequently determining the actual location of \( R_d \). Furthermore, to accurately identify \( R_{target} \), it is necessary to eliminate \( R_{d,noise} \) from \( R_d \). To address this issue, the proposed overall architecture is presented in Figure \ref{fig:flowchart}.

\subsection{Vehicle inference using 2D radar PCD and images}\label{sec:spatial}
In general, static objects affecting urban alley navigation primarily include buildings and parked vehicles. However, vehicles are relatively small and non-linear structures, making it challenging to infer their shapes in the $R$ because it is too sparse. The acquisition path of $R_d$ is highly dependent on the position and pose of the reflectors, which introduces spatial dependency. Misinterpretation of the spatial state of vehicles, which act as reflectors, could lead to erroneous reflection path estimations. 

To tackle this limitation, additional information observed from another sensor is required, therefore, $I$ has been utilized to enhance spatial interpretation. Although the distance information in $I$ is less accurate than that in $R$, it provides object shapes and class information, enabling the estimation of vehicle's spatial states.

\subsubsection{Depth estimation from front image}\label{sec:spatial-2}
To estimate $D$ from $I$, we employ a depth estimation model $f_{depth}(\cdot)$, defined as follows: 
\begin{align}
        D &= f_{depth}(I).
\end{align}

% \textcolor{red}{
% 이때 $D(u_p, v_p)$는 기존에 학습된 데이터를 기반으로 예측된 metric distance이기 때문에, 이를 본 연구의 실험환경과 맞춰줘야한다. 이를 위해, xxxx 방법을 기반으로 front cam image를 depth point cloud $P_D$로 변환한다, as formulated follows:
% \begin{align}
%         P_D &= \{D(u_p, v_p) \cdot K^{-1}\cdot [u_p, v_p, 1] ^T\}_{p=1}^N, \\
%         &=\{(X_p, Y_p, Z_p)\}_{p=1}^N, 
% \end{align}
% where $K$는 camera intrinsic matrix, $N$ 은 전체 이미지 pixel 수다. 
% }
% Since $D$ represents metric distance predictions based on pre-trained data, it needs to be calibrated to match the camera settings by transforming the depth map into depth point cloud $P_D=\{(X_p, Y_p, Z_p)\}_{p=1}^N$ using the camera intrinsic matrix $K$ and an unprojection function $f_{unproj}(\cdot, K)$, formulated as follows: 
% \begin{align}
%         P_D &= f_{unproj}(D, K), 
% \end{align}
% where $X_p, Y_p, Z_p$ denote the 3D coordinates of unprojected depth point cloud at pixel $p$, and $N$ represents the total number of pixels in the image. 

Since $D$ represents metric distance predictions based on pre-trained data, it needs to be calibrated to match the camera settings by transforming the depth map into depth point cloud $P_D$ using the camera intrinsic matrix $K$ and an unprojection function $f_{unproj}(\cdot, K)$, formulated as follows: 
\begin{align}
        P_D &= f_{unproj}(D, K) \\
        &=\{p \mid p= (p_x, p_y, p_z)\}.
\end{align}
where $p_x, p_y, p_z$ denote the 3D coordinates of unprojected depth point $p$.

\subsubsection{Vehicle Region-of-Interest (RoI) extraction}\label{sec:spatial-2}
% Since the distance information in $P_D$ is less reliable than that in $R_s$, it must be refined using corresponding static radar points. However, directly matching all $P_D$ points with all $R_s$ points would fail to accurately correct individual objects due to object-wise variation in depth estimation errors. Therefore, object-wise segmentation of $P_D$ is performed by several steps. First, vehicle mask $M_{vehicle}$ is obtained by using a segmentation model $f_{seg}(\cdot)$, as formulated follows:
% \begin{align}
%         M_{vehicle}=f_{seg}(I),
% \end{align}
% and vehicle mask at pixel $M_{vehicle}(c_i)$ is 1 if $f_{seg}(c_i)$ is vehicle and 0 when $f_{seg}(c_i)$ is not vehicle. 

Since the distance information in \( P_D \) is less reliable than that in \( R_s \), it must be refined using corresponding static radar points. However, directly matching all points in \( P_D \) with those in \( R_s \) would fail to accurately correct individual objects due to object-wise variations in depth estimation errors. To address this, object-wise segmentation of \( P_D \) is performed through several steps. First, a vehicle mask, \( M_{vehicle} \), is obtained using a segmentation model \( f_{seg}(\cdot) \), formulated as follows:
\begin{align}
    M_{vehicle} = f_{seg}(I),
\end{align}
where the vehicle mask at pixel \( c \), denoted as \( M_{vehicle}(c) \), is 1 if \( f_{seg}(c) \) identifies a vehicle, and 0 otherwise.

Then, the segmented pixel coordinate set corresponding to vehicles $C_{vehicle}$ is denoted as follows:
\begin{align}
    C_{vehicle} &= \{ c\mid M_{vehicle}(c) = 1 \}.
\end{align}

To remove low-density noise regions, DBSCAN \cite{ester1996density} is applied to $P_{D, vehicle}=f_{unproj}(D(C_{vehicle}),K)$ forming a set of clusters in depth PCD $\mathcal V= \{ V_1,\ \dots , V_m\}$, where $V_m$ is a vehicle cluster, $m$ is total number of vehicle clusters.

\subsubsection{Rough estimation of vehicle spatial state}\label{sec:spatial-2}
Since $I$ is captured from the ego vehicle's perspective, it can only observe one or two surfaces of a vehicle (e.g., front + side, front-only, or side-only). Consequently, $V$ also contains at most two surfaces of the vehicles. These surfaces must be distinguished and optimally aligned to a standard vehicle size, defined as the product of the vehicle's width $W$ and vehicle's length $L$.

The initial step in this process involves projecting \( V \) onto the XY-plane as the ground plane. The clusters are then divided into two groups based on the median Y-axis to determine whether a horizontal surface is present. Subsequently, piecewise linear regression is applied to each segment to extract the corresponding vehicle surface edges. 

If at least one horizontal line with a slope $\theta$ less than 45\textdegree{} is found, the vehicle is likely positioned in front of the ego vehicle, meaning that the horizontal line corresponds to the closest surface of the vehicle. In contrast, if only vertical lines are detected, the vehicle is likely located at a similar Y-coordinates as the ego vehicle, meaning the detected vertical line corresponds to the vehicle's side surface. 

Using this classification, the rough center point $v_c'=(x_c', y_c')$ of the vehicle is computed as follows, which serves as an initial estimate before refinement to obtain the final center point $v_c$: 
\begin{align}
    \label{eq:vehicle}
        v_c' &=
        \begin{aligned}
            \begin{cases} 
                (x_{min} + \frac{1}{2}W, y_{min} + \frac{1}{2} L), &  \textstyle if \ exist \ \theta < 45\degree \\ 
                (x_{min} + \frac{1}{2}W, y_{max} -  \frac{1}{2}L), & \text{otherwise} 
            \end{cases},
        \end{aligned}    
\end{align}
where, $x_{min}, y_{min}, x_{max}, y_{max}$ refers to the maximum and minimum values of the x and y coordinates among all points within the $V$. Using this estimation, a rough bounding box $B'= (v_c', W, H)$ is placed around the vehicle to approximate its spatial positioning.

\subsubsection{Refinement of vehicle position using radar PCD}\label{sec:spatial-2}
$B'$ estimated based on $P_{D,vehicle}$ contains inaccuracies in distance and scale, necessitating further refinement using $R_s$. Since radar reflections typically occur on the vehicle's outer surface, despite the sparsity and noise of $R_s$, it still provides valuable surface information. 

To refine $B'$, a similarity evaluation is performed between $B'$ and each radar point in $R_s$ based on distance. Only $R_s$ within $[(x_c'\pm W,y_c'\pm L)]$ are considered, denotes as $R_{near}$. The similarity score is computed by evaluating the minimum distance $d(B',r_{near})$ between each point $r_{near}\in R_{near}$ and edges of $B'$. If $d(B',r_{near})$ is below a threshold $\tau$, the score increases. Subsequently, a grid search within a search range of \( \pm \delta \) is performed to determine the optimal bounding box \( B=(v_c, W, H) \) that maximizes the similarity score, formulated as follows: 
\begin{align} 
    B = \arg \max_{B'}|\{r_{near}\mid(d(B',r_{near})\leq\tau )\}|. 
\end{align} 
% where, $\gamma$ represents the number of search grid steps. 

\subsubsection{Decision of final spatial information}\label{sec:spatial-2}
In certain frames, the segmentation model $f_{seg}(\cdot)$ may fail to detect vehicles, or the estimated depth map $P_{D,vehicle}$ may exhibit severe outliers. In such cases, the absence of a reflector may result in misinterpreting $R_d$ as a direct signal, leading to errors in reflection path estimation and ultimately degrading localization accuracy. To stabilize spatial information, a frame-averaging approach is applied over five frames, including the current frame and the previous four frames.

After a pedestrian is detected in the NLoS region, the presence of the pedestrian may interfere with sensor measurements, leading to inaccuracies in spatial information. To mitigate this, the spatial information is fixed from the moment of the initial pedestrian detection, using the previously predicted spatial information.

The final accumulated bounding box  The final accumulated bounding box $\bar{B_{t}}=(\bar{v}_c, W, H)$ at frame t, as the final spatial information estimation $S$, is defined as follows: 
\begin{align} 
    S = \bar{B_{t}} = 
    \begin{cases} 
        \text{avg}(\sum_{i=1}^{t}B_{t}), & \text{if } t < 5 \\
        \text{avg}( \sum_{i=t-4}^{t} {B_t}), & \text{if } t \geq 5
    \end{cases}.
\end{align}
% To stabilize spatial information, a frame-averaging approach is applied over five frames including current frame and previous four frames. %\textcolor{red}{NLoS 영역에 pedestrian이 감지된 후에는, pedestrian의 센서 간섭으로 인해 }

% 공간정보를 부정확하게 만들 수 있으므로 최초 pedestrian 감지 시점부터는 직전에 예측한 공간정보로 고정한다.}  The final accumulated bounding box $\bar{B'_{t}}$ at frame t, as the final spatial information estimation $S$, is defined as follows: 

% \begin{align} 
%     \bar{{B'}}_{t} = 
%     \begin{cases} 
%         \frac{1}{5}\sum_{i=1}^{t}{{B'}}_{t}, & \text{if } t < 5 \\
%         \frac{1}{5} \sum_{i=t-4}^{t} {B'}, & \text{if } t \geq 5 
%     \end{cases} 
% \end{align}

\subsection{Estimation of target location}\label{sec:spatial}
\subsubsection{Integration with spatial information}\label{sec:spatial-2}
Using $S$ obtained from $R_s$ interpretation, the reflection path of mmWave signals can be estimated via ray tracing, enabling the localization of NLoS targets. 

Since mmWave signals exhibit strong reflectivity, they undergo mirror-like reflections upon striking walls. Thus, the reflection paths of $R_d$ points are estimated using ray tracing. 
For each dynamic points $r_i \in R_d$, a line segment is drawn from the radar origin $O$ to $r_i$. The first collision point $q_i$ with the spatial structure is computed. Given that the structure follows a linear equation $y_i=\alpha x_i + \beta$, the mirror-reflected x, y coordinates of $r_i'$, denoted as $r'_{x,i}$ and $r'_{y,i}$ is computed as follows:  
\begin{align}
    r'_{x,i} &= \frac{2 \alpha (r_{y,i} - \beta) + r_{x,i}}{\alpha^2 + 1} - r_{x,i}\ ,
\end{align}
\begin{align}
    r'_{y,i} &= 2 \left(\alpha \frac{(r'_{x,i} + r_{x,i})}{2} + \beta \right) - r_{y,i} \ ,
\end{align}
where $r_{x,i}$ and $r_{y,i}$ denotes the x, y coordinates of $r_i$ respectively.

This process is iteratively repeated for $r_i'$, treating $q_i$ as a new virtual radar origin $O$. The process continues until no further reflections occur, yielding the final corrected target location $R_d'$.

\subsubsection{Final target position determination }\label{sec:spatial-2}
DBSCAN clustering algorithm is applied to $R_d'$ to remove outlier noise, retaining only valid object clusters. Any cluster too close to the spatial structures is considered physically implausible and removed. Moreover, valid targets in our environment should generate both direct and reflected signals. If a cluster contains only direct rays, it is likely a ghost reflection and is thus filtered out. After all filtering steps, the remaining cluster set constitutes $R_{target}$, where the centroid of cluster represents the predicted object location $X_{pred}$.

\section{Dataset}\label{sec:Datasets}
To validate the proposed method, a test environment measuring 53.5 $m$ $\times$ 33.5 $m$ was specifically designed and constructed. A data acquisition vehicle was set up to collect experimental data. The test site was purposefully designed and built to evaluate the detection and localization of non-visible objects that suddenly appear between vehicles. To facilitate the observation of NLoS scenarios, a 12-megapixel camera with a fisheye lens providing a 185\textdegree{} Field-of-View (FoV) was installed at a height of 7 $m$. 

For data acquisition, the vehicle was equipped with 77 $GHz$ mmWave corner radar mounted on the lower right bumper to observe the NLoS region, while a front camera was installed at the front mirror position to monitor the LoS region. To simulate high-risk traffic scenarios involving potential collisions with NLoS objects, the vehicle was positioned 7 $m$ from the intersection center during data acquisition.

Further details regarding the test site and data acquisition vehicle can be found in Jeon \textit{et al.} \cite{jeon2024non}. The dataset scenario was configured by varying the presence, movement direction, and number of pedestrians in both LoS and NLoS regions while keeping two parked vehicles stationary.

% 제안한 방법의 검증을 위해 53.5 $m$ $\times$ 33.5 $m$의 세트장을 직접 설계, 건축하고 데이터 취득용 차량을 셋업하여 데이터를 취득했다. 특히 세트장은 차량 사이에서 갑작스럽게 등장하는 비가시 객체의 탐지 및 localization 방법의 검증을 목적으로 설계, 건축되었기 때문에, NLoS 영역에서의 상황 관측을 위해 7 $m$ 상공에 185\textdegree{}의 시야각을 갖는 fisheye lens가 부착된 12 megapixel 카메라를 설치하였다. 데이터 취득 차량의 경우 NLoS 영역 관측을 위해 2대의 77$GHz$ mmWave radar를 우측 하단 범퍼에 부착하고, LoS 영역의 상황 관측을 위해 전방 룸미러 위치에 front cam을 설치했다. 데이터 취득시에는 NLoS 객체로 인한 사고 가능성이 높은 상황을 고려하여 교차로 중심에서부터 7 $m$ 위치에 차를 배치하였다. 세트장 및 데이터 취득 차량에 대한 더 자세한 설명은 Jeon \textit{et. al.} \cite{jeon2024non}에 기술되어 있다. 데이터셋 시나리오 구성의 경우 두대의 차량을 주차해둔 상태에서 LoS 및 NLoS 영역에 보행자 존재 유무 및 움직임의 방향, 객체 숫자를 달리하여 다양한 상황들에 대해 취득했다, as depicted in Figure xxx.

% \subsection{Vehicle for data acquisition}
% For data collection, various sensors were mounted on an SUV, taking into account their installation positions on an actual vehicle. First, two 77 GHz mmWave radars were installed for detecting NLoS objects. Additionally, a front camera was mounted on the vehicle’s interior rearview mirror to confirm the NLoS or LoS status of objects, and a LiDAR system was installed for sensor calibration. Data from all sensors were acquired at 100 $ms$ intervals, and ROS Noetic was utilized to ensure synchronization of the data collection. Only the MIMO radar PCD was utilized in this paper.

\section{Experimental Results}\label{sec:Experiments}
The validation of the proposed methodology is conducted under three experimental scenarios. 
\textit{SA} represents a scenario where a single pedestrian darted from the NLoS region between two parked vehicles. Although simple, this is one of the most frequently occurring real-world situations. 

\begin{figure*}
    \centering
    \includegraphics[width=\linewidth]{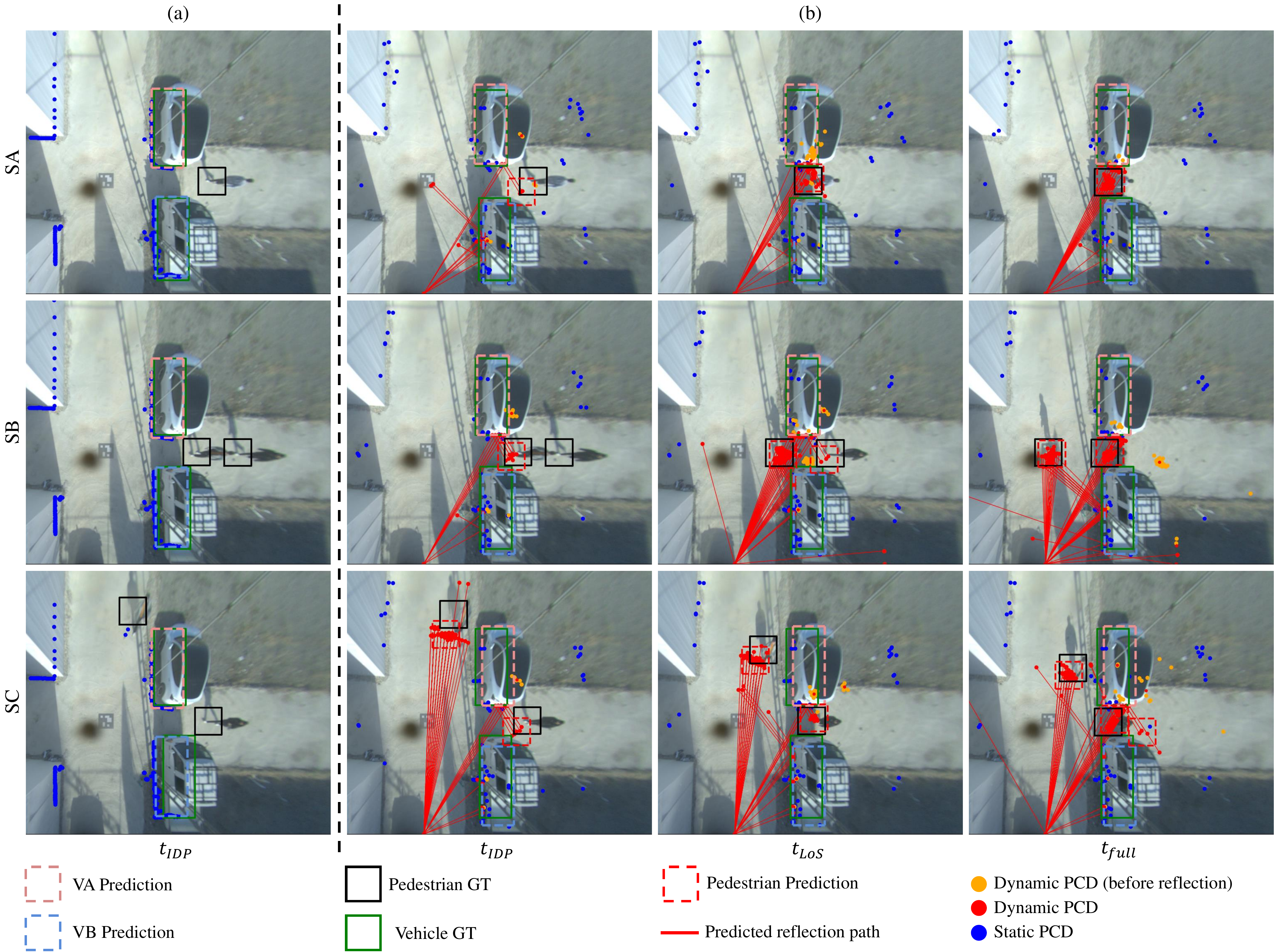}
    \vspace{-0.7cm}
    \caption{Experiment result of our method. (a) Spatial inference results of the LiDAR-based method.
(b) Spatial inference results of the radar-based method and the temporal evolution of the pedestrian localization model. (Left) the moment at Initial Detected Position (IDP). (Middle) the moment when a part of the pedestrian's body is first captured by the front camera. (Right) the moment when the entire body appears in the LoS region.}
    \label{fig:quanres}
\end{figure*}

The remaining two scenarios introduce increased complexity by involving two pedestrians. \textit{SB} represents a scenario where two pedestrians sequentially appear in the NLoS region between parked vehicles. Finally, \textit{SC} is similar to \textit{SA} in that one pedestrian darts out from the NLoS region between parked vehicles, but simultaneously, another pedestrian approaches the ego-vehicle from the LoS region. Additionally, in every scenario, two parked vehicles are present: the front parked vehicle is denoted as \textit{VA}, and the rear parked vehicle is denoted as \textit{VB}.

% The remaining two scenarios introduce increased complexity by involving two pedestrians. \textit{SB} represents a scenario where two pedestrians sequentially appear in the NLoS region between parked vehicles. Finally, \textit{SC} is similar to \textit{SA} in that one pedestrian darted from the NLoS region between parked vehicles, but simultaneously, another pedestrian approaches the ego-vehicle from the LoS region. Also, for every scenarios, 주차된 vehicle이 두 대 존재하며, 앞에 주차된 vehicle을 \textit{VA} 뒤에 주차된 vehicle을 \textit{VB}라고 칭한다. 

In this paper, YOLOv8 \cite{Jocher_Ultralytics_YOLO_2023} was utilized as $f_{seg}(\cdot)$ to extract object contours, and for depth estimation, Depth Anthing V2\cite{yang2025depth} is utilized as $f_{depth}(\cdot)$ to infer the rough spatial distance from monocular image.

\subsection{Evaluation of spatial inference model}
Since the NLoS pedestrian location is inferred through ray tracing based on the estimated spatial information, the accuracy of the spatial inference model is a critical factor in NLoS localization performance. As similar to Equation \ref{eq:3}, to evaluate the accuracy of the predicted position, the Euclidean distance $E$ between the predicted location $(x_{pred}, y_{pred})$ and the ground truth $(x_{gt}, y_{gt})$ is used as an evaluation metric, defined as follows:
\begin{align}
    E = \sqrt{(x_{pred} - x_{gt})^2 + (y_{pred} - y_{gt})^2}.
\end{align}

It is considered as correct estimation if $E$ is larger than threshold value, which is emphirically set 0.2 in this paper.

To evaluate the spatial inference model, $\bar{v}_c$ is compared with the ground truth center point obtained from BEV camera, which is denoted as $v_{c, gt}$. Additionally, to evaluate sparsity of $R$ on model performance, $\bar{v}_c$ is compared with $\bar{v}_{c,lidar}$, which follows the same processing pipeline bue replaces $R$ with a denser LiDAR PCD. The qualitative results of spatial inference model is depicted in Figure \ref{fig:quanres}(a).

First, for \textit{VA}, which is largely visible in $I$, the differences between the LiDAR-based case and the proposed method were within 0.1 $m$ in \textit{SA} and \textit{SB}, confirming that precise refinement was achieved. This result validates that the proposed method effectively infers spatial information. However, in \textit{SC}, the discrepancy increased due to the presence of a pedestrian in the LoS region near the vehicle. This resulted in static observation points unrelated to vehicles appearing in $R_s$, leading to an increased number of non-vehicle elements. This issue can be addressed by filtering out obstacles in the LoS region when selecting candidate reflectors $R_{reflector}$.

Furthermore, in all scenarios, the spatial information inferred using LiDAR PCD exhibited an average deviation of approximately 0.19 $m$ from $v_{c,gt}$. This discrepancy is attributed to cumulative calibration errors across sensor data. It is likely caused by the absence of precise landmark-based calibration in the test bed used for data collection. However, since the error remains consistent across scenarios, it can be eliminated through precise sensor calibration, confirming the validity of the proposed model's performance.

Comparing cases where the input to the proposed model is radar PCD versus LiDAR PCD, the discrepancy between the ground truth of \textit{VA} and the estimated position was up to 76\% smaller than that of \textit{VB}. This indicates that incorporating more camera-derived information enhances the accuracy of spatial inference, demonstrating the benefits of multimodal fusion. Additionally, the errors observed when using LiDAR PCD were consistently lower than those observed when using radar PCD, indicating that higher PCD density leads to more accurate spatial inference. The overall experiment results are described in Table \ref{tab:QuanResLocal1}.

\begin{table}[t!]
\centering
\caption{Evaluation of Spatial Inference Model.}
\label{tab:QuanResLocal1}
\tiny
\vspace{-0.2cm}
\resizebox{\columnwidth}{!}{
    \begin{tabular}{c|c|cc|c}
            % \midrule[0.9pt]
            \hline
            \multirow{2}{*}{Scenarios}  & Target & \multicolumn{3}{c}{Distance from $v_{c,gt}$ [\textbf{m}]}   \\
            \cline{3-5}
             &  Vehicles & Radar & LiDAR & Diff. \\  \hline\hline
            
            \multirow{3}{*}{\textit{SA}}& \textit{VA} & 0.26 & 0.17 & 0.09 \\ 
            & \textit{VB} &  0.41 & 0.21 & 0.20  \\
            \cline{2-5}
            &Avg.&0.34&0.19&0.15\\
            \hline

            \multirow{3}{*}{\textit{SB}}& \textit{VA} & 0.13 & 0.19 & 0.06 \\ 
            & \textit{VB} & 0.44 & 0.19 & 0.25 \\
            \cline{2-5}
            & Avg. &0.29&0.19&0.16\\
            \hline

            \multirow{3}{*}{\textit{SC}}& \textit{VA} & 0.43 & 0.17 & 0.26 \\ 
            & \textit{VB} & 0.60 & 0.19 & 0.41 \\
            \cline{2-5}
            & Avg. &0.52&0.18&0.34\\
            \hline
            
        \end{tabular}
    }
\end{table}

% 먼저, $I$에서 많은 부분이 관측된 \textit{VA}의 경우, \textit{SA}와 \textit{SB}에서 LiDAR case와 0.1 $m$ 이내의 차이를 보이며, 정밀한 refinement가 이루어졌음을 확인했으며, 이는 제안한 방법이 효과적으로 공간정보를 추론했음을 의미한다. 다만 \textit{SC} 에서는 차이가 증가하였는데, 이는 차량의 주변의 LoS 영역에 pedestrian이 존재하여, 이에 대한 관측값 일부가 static 하여, 차량 이외의 $R_s$가 많이 존재하기 때문이다. 이는 추후에 LoS 영역의 obstacle들은 탐색 후 $R_{reflector}$의 후보군에서 제외하여 해결할 수 있다. 

% 또한, 모든 시나리오에서 LiDAR PCD를 입력으로 추론한 공간정보는 $v_{c,gt}$와 약 0.19 $m$의 차이를 보였다. 이는 각 센서 데이터의 calibration 과정에서 발생한 오차가 누적되어 발생한 것으로, 데이터 취득을 위한 test bed에는 landmark가 부재하여 이를 정밀하게 교정하지 못했기 때문으로 추정된다. 하지만 정밀한 sensor calibration이 동반된다면 해당 오차는 제거 가능하며, 모든 시나리오에 대해 오차가 일정하기 때문에 이를 감안한다면 제안한 모델의 성능이 유효함을 확인할 수 있다.

% 제안한 모델의 input이 radar PCD인 경우와 LiDAR PCD 인 경우를 차이를 비교하면, 먼저 전반적으로 $I$에서 더 많은 부분을 확인할 수 있는 \textit{VA}의 ground truth와의 오차가 \textit{VB}의 ground truth와의 오차에 비해 최대 76\% 작게 나타났다. 이는 camera로부터 취득한 정보가 많이 반영될 수록 정확한 공간정보 추론이 가능함을 의미하여, multimodality 활용의 효용성을 확인할 수 있다. 이외에도 LiDAR PCD를 활용했을 때의 오차가 radar PCD를 활용했을때의 오차보다 모두 작게 나타나는데, 이는 PCD의 density가 높을수록 더 정확한 공간정보 추론이 가능함을 의미한다.

\subsection{Evaluation of Pedestrian Localization Model}
To prevent pedestrian accidents in road environments with roadside-parked vehicles, it is crucial to detect pedestrians as quickly and accurately as possible using a ray tracing model within the estimated spatial information $S$. Therefore, the pedestrian localization model is evaluated in terms of detection range and accuracy.
% 갓길 주차된 차도에서 보행자 사고를 예방하기 위해서는, 추정한 공간 $S$에서 ray tracing 모델을 활용했을 때 최대한 빠르고 정확하게 pedestrian을 탐지해 내는 것이 관건이다. 따라서 Pedestrian localization model을 detection range와 accuracy 두 가지 측면에서 비교한다. 

Localization is considered successful if the Intersection over Union (IoU) between the predicted pedestrian 1.7 $m$ \(\times\) 1.7 $m$ bounding box \( B_{ped} \) with center point $X_{pred}$ and the ground truth bounding box \( B_{ped,gt} \) is at least 0.2:  
\begin{align}
    \text{IoU} &= \frac{|B_{ped}\cap B_{ped,gt}|}{|B_{ped} \cup B_{ped,gt}|}.
\end{align}

To assess the effective detection range of the pedestrian localization model, the x-axis distance between the ego-vehicle's origin \( O \) and the pedestrian's ground truth position at the first NLoS detection frame is analyzed as the Initial-Detected-Position (IDP), when the IoU exceeds 0.2.
\begin{align}
    \text{IDP} = | x_{GT} - O |, \quad \text{when } \text{IoU} > 0.2,
\end{align}
where, $x_{GT}$ is the $x$-coordinate of $X_{GT}$.

Furthermore, the time interval \( t_{\text{IDP}} \) from the initial detection to the moment when the pedestrian becomes fully visible in the LoS region at \( t_{\text{full}} \) is examined. This interval is defined as Time-To-Appearance (TTA) as formulated follows:  
% To evaluate the 유효 detection range of the pedestrian localization model, IoU가 0.2 보다 높은 첫 NLoS 객체 탐지 frame (Initial-Detected-Position) ego-vehicle의 원점 $O$와 pedestrian 사이의 x축 거리, 처음 등장 시점 $t_{IDP}$로부터 pedestrian이 LoS 영역에 완전 등장까지의 시간인 $t_{full}$을 비교하여 Time-To-Appearance (TTA) 를 평가한다. 
\begin{align}
    TTA = t_{full} - t_{IDP}.
\end{align}

The initial appearance $t_{LoS}$ and full appearance $t_{full}$ are determined based on the front-camera image captured from the driver’s perspective. The initial appearance is defined as the first moment when any part of the pedestrian’s body becomes visible in the front-camera image, while the full appearance is when the entire pedestrian body becomes visible. When multiple pedestrians are detected, they are numbered sequentially, starting from Ped.1 to Ped.2 according to the order of detection.
% 첫 등장과 완전 등장은 운전자의 시점에서 취득된 이미지인 frontcam image 상에서 처음 신체의 일부가 보였을 때와, 신체의 모든 부분이 LoS 영역에 존재할 때를 기준으로 한다. Pedestrian이 2명 이상일 때 먼저 탐지가 되는 객체부터 1번부터 넘버링한다. 

\begin{table}[t!]
\centering
\caption{Evaluation of Pedestrian Localization Model.}
\label{tab:QuanResLocal2}
\vspace{-0.2cm}
\resizebox{\columnwidth}{!}{
    \begin{tabular}{c|c|c|c|c}
            % \midrule[0.9pt]
            \hline
            \multirow{2}{*}{Scenarios}  & \multicolumn{2}{c|}{Initial detection} & Accuracy $\uparrow$  & AE  $\downarrow$  \\
            \cline{2-3}
             &  TTA [\textbf{s}] $\uparrow$ & IDP [\textbf{m}] $\uparrow$ & [\textbf{\%}] & [\textbf{m}] \\  \hline\hline
            
            \textit{SA}& 1.9 & 6.9 & 90.48 & 0.38 \\ 
            \hline

            \textit{SB}& 0.95 & 5.99 & 89.19 & 0.31 \\ 
            \hline

            \textit{SC}& 1.4 & 6.5 & 81.25 & 0.58 \\ 
            \hline
            
        \end{tabular}
    }
\end{table}

To assess the accuracy of the pedestrian localization model, the ratio of frames in which the IoU between the predicted and ground truth positions exceeds 0.2 is calculated, from the IDP to the full LoS appearance, as defined follows:
\[
\text{Accuracy} = \frac{N_{\text{IoU} > 0.2}}{N_{\text{total}}},
\]
where \( N_{\text{IoU} > 0.2} \) is the number of frames where the detected pedestrian bounding box satisfies over 0.2 IoU, and \( N_{\text{total}} \) is the total number of frames from $t_{IDP}$ to $t_{full}$.

Furthermore, since minimizing the distance between the predicted position $X_{pred}$ and the ground truth $X_{GT}$  is crucial, the Absolute Error (AE) is evaluated across all frames where the IoU surpasses 0.2 within this interval, following Equation \ref{eq:3}.
% To evaluate accuracy of the pedestrian localization model, IDP로부터 모든 pedestrian이 LoS 영역에 완전히 등장하는 시점까지 IoU가 0.2 보다 높은 탐지의 비율을 구한다. $X_{pred}$와 $X_{GT}$간의 거리를 최소화했는지 평가해야 하므로, IDP로부터 완전 등장 시점까지 IoU가 0.2 보다 높은 탐지에서 ground truth 값까지의 Absolute Error (AE) 를 구한다, based on Equation \ref{eq:3}. 
 The quantitative performance of the model evaluated based on these metrics is presented in Table \ref{tab:QuanResLocal2}, and qualitative results are depicted in Figure \ref{fig:quanres}(b).

In \textit{SA}, the pedestrian was first detected 1.4 seconds before appearing in the front-camera image and 1.9 seconds before fully darting out into the LoS region. In other words, the pedestrian was detected 1.9 seconds before full visibility. IDP of the pedestrian was 6.9 $m$. The accuracy was 90.48\%, and the AE for detected pedestrians was 0.38 $m$.  

In \textit{SB}, Ped.1, who was walking quickly, was first detected 0.5 seconds before initially appearing and 0.9 seconds before fully visible. IDP of Ped.1 was 5.94 $m$. Despite being occluded by Ped.1, Ped.2 was also detected 0.7 seconds before its first appearance in the LoS region and 1.0 second before full visibility. IDP of Ped.2 was 6.04 $m$ from \( O \). The overall detection accuracy from Ped.1’s $t_{\textit{IDP}}$ to Ped.2’s $t_{full}$ was 89.19\%, and the AE was 0.31 $m$.  

In \textit{SC}, Ped.1 was initially in the LoS region, while Ped.2 started in the NLoS region. From the IDP of Ped.2 to its first LoS appearance was 1.0 second, and the time until full visibility was 1.4 seconds. IDP of Ped.2 from \( O \) was 6.5 $m$. The overall detection accuracy for Ped.2's $t_{\textit{IDP}}$ to Ped.2's $t_{full}$ was 81.25\%, and the AE was 0.58 $m$.  

Across all scenarios, the IDP remained within 6 to 7 $m$, indicating a detection range limitation caused by the distance between parked vehicles in the experimental setup. Since \textit{VA} is assumed to be the primary reflector, the radar signal detects NLoS objects through the gap between \textit{VA} and \textit{VB}. Consequently, if the ego-vehicle is positioned further ahead, the FoV restriction from \textit{VB} is reduced, enabling detection over a greater range. Conversely, if the ego-vehicle is further behind, the FoV restriction from \textit{VB} becomes more severe, limiting detection range. However, a rearward position provides a longer reaction time before a potential collision, while a forward position increases detection range, thereby enhancing accident prevention.  

\textit{SB} and \textit{SC} exhibited lower detection accuracy compared to \textit{SA}. The primary reason for this decrease is the presence of multipath reflections from pedestrians in the LoS region, which resulted in ghost targets. Some of these ghost targets appeared similar to direct signals, making them difficult to filter using our clutter filtering algorithm. However, such signals can be easily filtered out using the front camera. Another contributing factor was that pedestrians in the LoS region obstructed the radar signal path used for detecting NLoS objects, directly interfering with detection.  

In \textit{SC}, the AE for detected pedestrians was higher than in \textit{SA} and \textit{SB}. This is because Ped.1, which was originally in the LoS region, was detected from a greater distance, causing its position to be outside the precise radar detection range. As a result, even though a direct radar signal was received, distance estimation errors occurred.

\section{Conclusions}\label{sec:Conclusion}
In this paper, we demonstrated that NLoS pedestrian localization is feasible by utilizing a camera assistant to obtain class information for points that cannot be extracted from radar PCD. Using this approach, we identified the locations of vehicles and inferred spatial information, enabling NLoS pedestrian localization.  

The proposed method achieved NLoS pedestrian detection within an average x-distance of 6 m from the ego-vehicle, demonstrating its capability to provide sufficient additional time for braking. Currently, this approach focuses solely on vehicle detection; however, future improvements could incorporate information on pedestrians and other LoS objects to enhance robustness against noise.  

The accuracy of the pedestrian localization model was 86.97\%. Among the detected frame, the absolute distance error was 0.42 $m$. The algorithm showed high reliability to help prevent accidents in real-world environments.

The scenario in which a pedestrian walks toward the vehicle within the LoS area exhibited the lowest accuracy; however, this is an unrealistic situation. In the most realistic scenario, where a pedestrian emerges from between parked vehicles, the model achieved a high accuracy of over 90\%.

% 이 논문에서 우리는 camera assistant를 이용하여 레이더 PCD에서는 얻을 수 없었던 point의 class 정보를 구하고, 이를 통하여 vehicle의 위치를 찾아, 추론한 공간정보 상에서 NLoS pedestrian localization 이 가능함을 보였다.  NLoS pedestrian 탐지 성능은 평균 ego-vehicle로부터 x distance 6 $m$ 이내에서 탐지하여, 실제 브레이크를 작동시키기에 충분한 추가 시간을 제공할 수 있음을 보였다. 현재는 이 방식에서 vehicle에 대한 탐지만 진행하지만, 추후에 pedestrian이나 다른 LoS 영역의 객체 정보도 포함하여 더욱 noise에 강건한 모델롭 발전시킬 수 있다.
% In this paper, we proposed a method for estimating the locations of multiple dynamic pedestrians in NLoS regions using 2D radar point cloud data based on mmWave radar. The method involves inferring spatial configurations, calculating reflection paths, and removing noise with clustering to effectively estimate the positions of NLoS pedestrians. Experimental results revealed that the proposed method could predict the positions of pedestrians in NLoS regions with an error of less than 0.5 $m$. Future research will aim to extend the approach to more diverse environments, which includes dynamic ego-vehicle scenarios.

\addtolength{\textheight}{-12cm}   % This command serves to balance the column lengths
                                  % on the last page of the document manually. It shortens
                                  % the textheight of the last page by a suitable amount.
                                  % This command does not take effect until the next page
                                  % so it should come on the page before the last. Make
                                  % sure that you do not shorten the textheight too much.

\bibliographystyle{IEEEtran}
\bibliography{reference,IEEEabrv}

\end{document}